\documentclass[conference]{IEEEtran}
\IEEEoverridecommandlockouts
\usepackage{cite}
\usepackage{amsmath,amssymb,amsfonts}
\usepackage{algorithmic}
\usepackage{graphicx}
\usepackage{textcomp}
\usepackage{xcolor}
\usepackage{hyperref}
\usepackage{array}
\usepackage{titlesec}
\usepackage{soul}
\usepackage{multirow}
\def\BibTeX{{\rm B\kern-.05em{\sc i\kern-.025em b}\kern-.08em
    T\kern-.1667em\lower.7ex\hbox{E}\kern-.125emX}}
\begin{document}

\makeatletter
\newcommand{\linebreakand}{%
  \end{@IEEEauthorhalign}
  \hfill\mbox{}\par
  \mbox{}\hfill\begin{@IEEEauthorhalign}
}
\makeatother

\title{D-SarcNet: A Dual-stream Deep Learning Framework for Automatic Analysis of Sarcomere Structures in Fluorescently Labeled hiPSC-CMs
\thanks{Corresponding author: hieu.ph@vinuni.edu.vn (Hieu Pham).}\\
}
\author{\IEEEauthorblockN{1\textsuperscript{st} Huyen Le}
\IEEEauthorblockA{\textit{VinUni-Illinois Smart Health Center} \\
\textit{College of Engineering \& Computer Science}\\
\textit{VinUniversity}, Hanoi, Vietnam\\ 
\texttt{huyen.lm@vinuni.edu.vn}}
\and
\IEEEauthorblockN{2\textsuperscript{nd} Khiet Dang}
\IEEEauthorblockA{\textit{VinUni-Illinois Smart Health Center} \\
\textit{VinUniversity}\\
Hanoi, Vietnam \\
\texttt{khiet.dtt@vinuni.edu.vn}}
\and
\IEEEauthorblockN{3\textsuperscript{rd} Nhung Nguyen}
\IEEEauthorblockA{\textit{College of Health Science} \\
\textit{VinUniversity}\\ Hanoi, Vietnam\\
\texttt{nhung.nt@vinuni.edu.vn}}
\linebreakand 
\IEEEauthorblockN{4\textsuperscript{th} Mai Tran}
\IEEEauthorblockA{\textit{College of Engineering \& Computer Science} \\
\textit{VinUni-Illinois Smart Health Center} \\
\textit{VinUniversity}, Hanoi, Vietnam\\
\texttt{mai.tt@vinuni.edu.vn}}

\and
\IEEEauthorblockN{5\textsuperscript{th} Hieu Pham}
\IEEEauthorblockA{\textit{VinUni-Illinois Smart Health Center} \\
\textit{College of Engineering \& Computer Science}\\
\textit{VinUniversity}, Hanoi, Vietnam\\
\texttt{hieu.ph@vinuni.edu.vn}}
}
\maketitle

\begin{abstract}
Human-induced pluripotent stem cell-derived cardiomyocytes (hiPSC-CMs) are a powerful tool in advancing cardiovascular research and clinical applications. The maturation of sarcomere organization in hiPSC-CMs is crucial, as it supports the contractile function and structural integrity of these cells. Traditional methods for assessing this maturation like manual annotation and feature extraction are labor-intensive, time-consuming, and unsuitable for high-throughput analysis. To address this, we propose D-SarcNet, a dual-stream deep learning framework that takes fluorescent hiPSC-CM single-cell images as input and outputs the stage of the sarcomere structural organization on a scale from 1.0 to 5.0. The framework also integrates Fast Fourier Transform (FFT), deep learning-generated local patterns, and gradient magnitude to capture detailed structural information at both global and local levels. Experiments on a publicly available dataset from the Allen Institute for Cell Science show that the proposed approach not only achieves a Spearman correlation of 0.868—marking a 3.7\% improvement over the previous state-of-the-art—but also significantly enhances other key performance metrics, including MSE, MAE, and $R^2$ score. Beyond establishing a new state-of-the-art in sarcomere structure assessment from hiPSC-CM images, our ablation studies highlight the significance of integrating global and local information to enhance deep learning networks’ ability to discern and learn vital visual features of sarcomere structure.
\end{abstract}


\section{Introduction}

Cardiovascular diseases (CVDs) are the primary cause of death worldwide, contributing to a substantial number of fatalities and disabilities \cite{DiCesare2024}. Thus, the need for accurate and reliable tools in cardiac research is essential. The complexity of cardiac physiology and the intricate mechanisms underlying CVDs require sophisticated models that can faithfully replicate human cardiac function. Human-induced pluripotent stem cell-derived cardiomyocytes (hiPSC-CMs), with their unlimited, personalized source, are emerging as a promising alternative for drug discovery, disease modeling, and regenerative and precision medicine in cardiovascular fields, which may help to address this global health issue.\cite{hnatiuk2021human, haneke2022progress}. 

Quantitative methods to assess the maturation of hiPSC-CMs are essential, as hiPSC-CMs are still significantly less mature than human adult cardiomyocytes \cite{ahmed2020brief}. Several approaches based on electrophysiological attributes, metabolism, and gene expression profiles have been employed. For example,  evaluating electrophysiological parameters, the TNNI3 to TNNI1 ratio, and the production of IK1 shows potential although involves technical challenges \cite{ahmed2020brief}. Moreover, transcriptome-based approaches, such as a gene regulatory system proposed by Uosaki \textit{et al.} \cite{uosaki2015transcriptional} and a relative expression orderings-based scoring method proposed by Chen \textit{et al.} \cite{chen2019qualitative}, have been suggested for more accurate maturation assessment. However, due to the scarcity of transcriptome data covering the full spectrum of human heart development, these techniques are limited to mouse PSC-CMs.


Another critical approach that supports existing methods for the functional assessment of hiPSC-CMs is the characterization of sarcomeres, their fundamental contractile units. Proper sarcomere development not only enhances the physiological relevance of hiPSC-CMs but also ensures their effectiveness in modeling cardiomyocyte behavior for both research and therapeutic purposes. The sarcomere consists of two primary contractile proteins, including myosin and actin, forming thick and thin polymeric filaments, respectively. Sarcomere organized architecture, including myosin and action interaction, ensures efficient and coordinated cardiac muscle contraction \cite{craig2004molecular}. One more essential component is the z-disc, which designates the lateral margins of sarcomeres and is crosslinked by proteins $\alpha$-actinin. This connection not only stabilizes the sarcomere's architecture but also plays a key role in force transmission across the muscle fiber \cite{knoll2011sarcomeric}. Many current studies visualize the sarcomere structures using anti-sarcomeric $\alpha$-actinin on the confocal microscopy to fluorescently label z-discs, providing a tool to study human sarcomere function non-invasively \cite{skorska2022monitoring}. However, biological and temporal variations lead to significant differences in sarcomere length, force, velocity, and structure, making challenges for the experts in viewing, analyzing, and quantifying these images \cite{telley2007sarcomere}.

Conventional analyses based on sarcomere images provide limited metrics and have low throughput due to the need for manual selection for regions of interest \cite{wakefield2022deep}. In addition, it is only suitable for assessing well-aligned sarcomeres from mature cardiac tissues. With the recent advancements in machine and deep learning, it is natural to employ a learning-based way to automatically quantify these images.

While some progress has been made in the field, current efforts on learning-based hiPSC-CMs quantitative analysis using fluorescent images are still in the infancy stage. Pasqualini \textit{et al.} \cite{pasqualini2015structural} established 11 metrics to quantify the progressive organization of sarcomeres in striated muscle cells throughout their development. Neural networks and tree-bagging algorithms were then applied to assess the maturity of sarcomere structure. However, because the datasets of hiPSC-CMs across different developmental stages are limited, the model was not trained on hiPSC-CMs but on primary cardiomyocytes from neonate rats (rpCMs). Gerbin \textit{et al.} \cite{gerbin2021cell} extracted and fed 11 cell features into linear regression to classify stages of sarcomere organization at the single-cell level. The model achieved a Spearman correlation of 0.67 and 0.63 on two testing sets. The feature engineering process was complicated, with six out of 11 features extracted from deep learning. SarcNet \cite{le2024sarcnet}, the current state-of-the-art framework, also demonstrated the ability of the ResNet-18 module to quantify sarcomere structural organization. However, the whole framework still relied on a prior feature extraction process, and the performance remains far from clinical applications. 

To address the above challenges, we propose D-SarcNet, a novel dual-stream deep learning framework that enables high-throughput and accurate quantification of sarcomere structure organization in hiPSC-CMs single-cell images. Specifically, the framework processes fluorescently labeled hiPSC-CM single-cell images as input and generates a continuous value ranging from 1.0 to 5.0, indicating the level of sarcomere structural organization for each image.  This approach significantly reduces the need for manual feature engineering by taking advantage of deep learning to extract high-level features directly and automatically. Remarkably, we propose three image-based representations: Fast Fourier Transform (FFT) Power image, local patterns, and gradient magnitude, to differentiate multiple patterns of $\alpha$-actinin-2 associated within the sarcomere such as fibers, puncta, and z-discs. We also design a dual-stream ConvNeXt-Swin Transformer architecture to enable the simultaneous acquisition of global and local information from the inputs. We summarize our contributions as follows:
\begin{enumerate}
\item We introduce a novel dual-stream deep learning framework to analyze sarcomere organizations in fluorescently labeled hiPSC-CMs single-cell images. Specifically, a ConvNeXt-Swin Transformer combined architecture is proposed to simultaneously acquire global and local patterns of the input image, allowing it to learn critical structural features and improve learning performance. 

\item We propose using the three image-based representations as inputs for the local features acquisition to provide the architecture with data based on frequency and sarcomere maturity, along with the magnitude of intensity variations for analysis.
\item We conduct extensive experiments and ablation studies to demonstrate the effectiveness of the proposed approach. Experimental results show that the proposed D-SarcNet outperforms previous state-of-the-art methods by a large margin. Our codes and pre-trained models are released at \href{https://github.com/vinuni-vishc/d-sarcnet}{https://github.com/vinuni-vishc/d-sarcnet} to encourage further studies on utilization of artificial intelligence (AI) to quantify sarcomere structures in hiPSC-CMs single-cell images.
\end{enumerate}
The rest of this paper is organized as follows. The problem setting of quantifying sarcomere structure organization and the details of the proposed framework are described in Section \ref{Methodology}. Details on experimental setup and results are presented in Section \ref{Experiments}. Finally, we conclude the paper by discussing the strengths and limitations in Section \ref{Discussion and Conclusion}. Supplementary materials can be found in the Appendix.

\section{Methodology} \label{Methodology}
This section discusses the proposed approach in details. We first formulate the problem as a regression task (Section \ref{Problem_formulation}). We then present an overview of the framework for predicting a continuous score of sarcomere structural organization on single-cell imaging of hiPSC-CMs (Section \ref{overall_framework}). Last, the framework architecture is described in Section \ref{framework_archi}.

\subsection{Problem Formulation} \label{Problem_formulation}
Given a set of \( N \) images of fluorescently labeled single-cell hiPSC-CMs, denoted as \( X = \{x_i\}_{i=1}^N \), with corresponding labels \( Y = \{y_i\}_{i=1}^N \), where \( y_i \) ranging from 1.0 to 5.0 represents the sarcomere structural maturity level for the image \( x_i \). We formulate this problem as a regression task where we aim to learn a function \(f_{\theta} : \mathcal{X} \rightarrow \mathcal{Y}\) that maps these images to their labels. This is done by training a deep learning model with parameters \( \theta \) to minimize the MSE loss over the training set. The MSE loss function is defined as
\begin{equation}
   \mathcal{L}(\theta) = \frac{1}{N} \sum_{i=1}^{N} |y_i - \hat{y}_i|^2,
\end{equation}
where \(y_i\) is the ground truth and \(\hat{y}_i\) is the predicted value.

\subsection{Overall Framework} \label{overall_framework}
\begin{figure*}
\centerline{\includegraphics[width=1\textwidth]{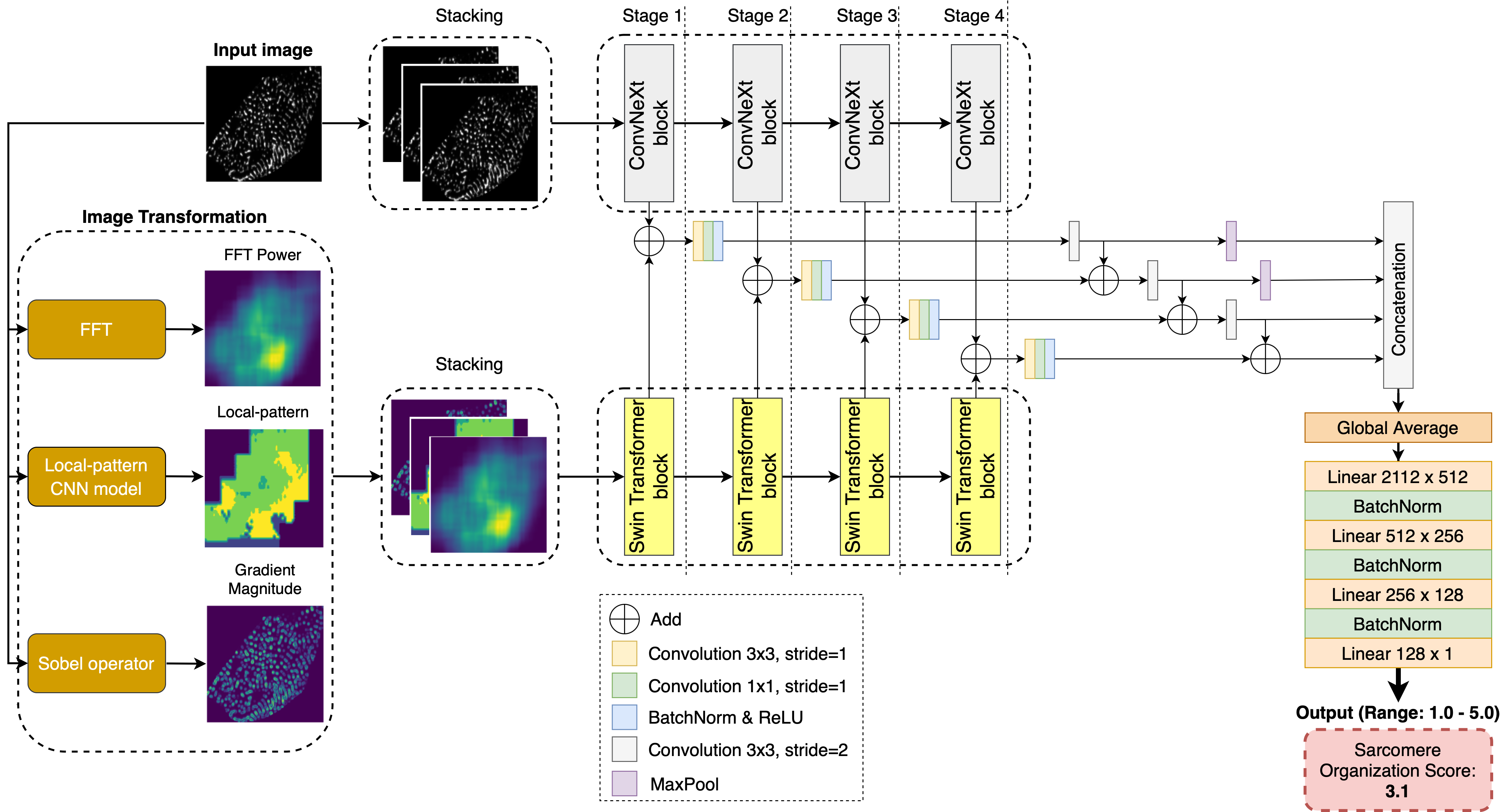}}
\caption{The D-SarcNet framework for scoring sarcomere structural organization in fluorescently labeled hiPSC-CM single-cell images consists of two primary streams. The first stream, ConvNeXt, processes raw images to extract global features directly from the original images. Simultaneously, the second stream, Swin Transformer, analyzes the corresponding three-channel image — created by stacking representations generated by FFT, the local pattern model, and the Sobel operator — to capture local features. A blocks-combined architecture then integrates feature maps from both streams across various scales to output a score of $\alpha$-actinin-2 pattern structure on a scale from 1.0 to 5.0.}

\label{fig_framework}
\end{figure*}
Fig. \ref{fig_framework} illustrates the overview of the proposed D-SarcNet architecture, which consists of two main streams: (1) The ConvNeXt model \cite{liu2022convnet} processes raw images to extract global features directly, (2) The Swin Transformer model \cite{liu2021swin,liu2022swin} analyzes the corresponding three-channel image created by stacking representations generated by FFT, the local-pattern model, and the Sobel operator to capture local features, focusing on the frequency domain, heterogeneity of $\alpha$-actinin-2 patterns, and the intensity variations. To integrate features from both streams at multiple scales, we propose a blocks-combined architecture that concatenates multi-scale feature maps into a single feature vector and then ultimately produces a score for the $\alpha$-actinin-2 pattern structure on a scale from 1.0 to 5.0. This is performed via a fully connected feedforward neural network for regression tasks, implemented with a series of linear layers, each followed by batch normalization.  

\subsection{Framework Architecture} \label{framework_archi}

\subsubsection{ConvNeXt Submodule} \label{convnext}
\begin{figure}
\centerline{\includegraphics[width=0.35\textwidth]{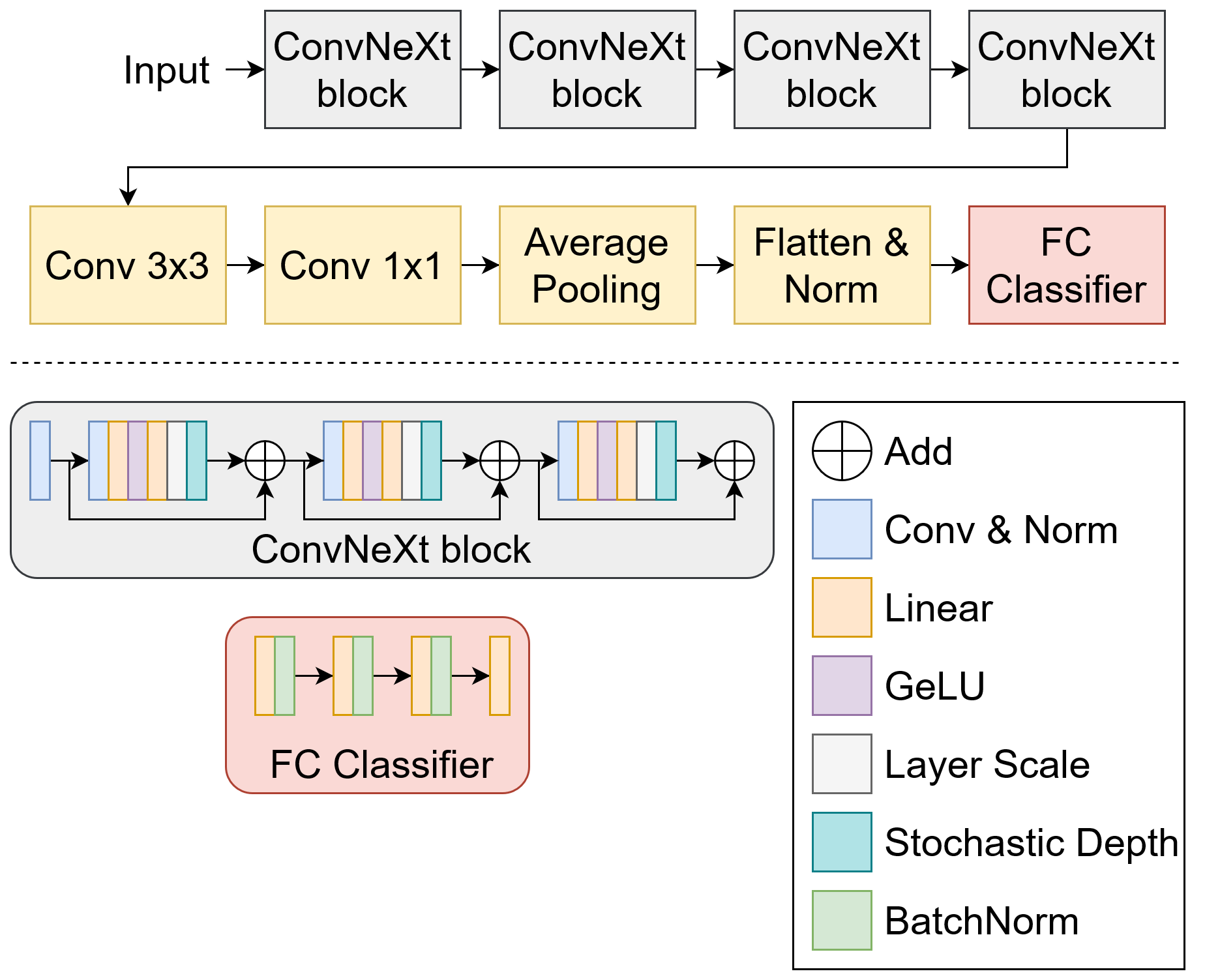}}
\caption{Illustration of the ConvNeXt framework. The whole framework has been applied to train the local-pattern model, and the ConvNeXt block has been utilized in the D-SarcNet model.}
\label{fig_convnext}
\end{figure}
ConvNeXt \cite{liu2022convnet} is a state-of-the-art module modernized from the standard ResNet \cite{he2016deep} based on the Vision Transformer \cite{dosovitskiy2020image}. Fig. \ref{fig_convnext} depicts the main backbone network consisting of four blocks, each utilizing an inverted bottleneck structure. Besides, with the usage of depth-wise convolution layers combined with $1 \times 1$ convolution, the system has led to the separation of spatial and channel mixing. In addition, convolution layers with global receptive fields ($7 \times 7$) are effective in extracting features on the macro scale. LayerScale is also applied to facilitate the convergence by initializing each channel weight with a small value as $\lambda_{i} = \varepsilon$. 

In this study, ConvNeXt has been applied to train the local-pattern model and the first stream of the D-SarcNet model. Regarding the local-pattern model, in the end, two more convolution layers with different kernels of $3 \times 3$ and $1 \times 1$ have been applied to gather diverse ranges of characteristic information via varied sizes of receptive fields. \cite{dong2022tc}. 

\subsubsection{Swin Transformer Submodule} \label{swin}
\begin{figure}
\centerline{\includegraphics[width=0.25\textwidth]{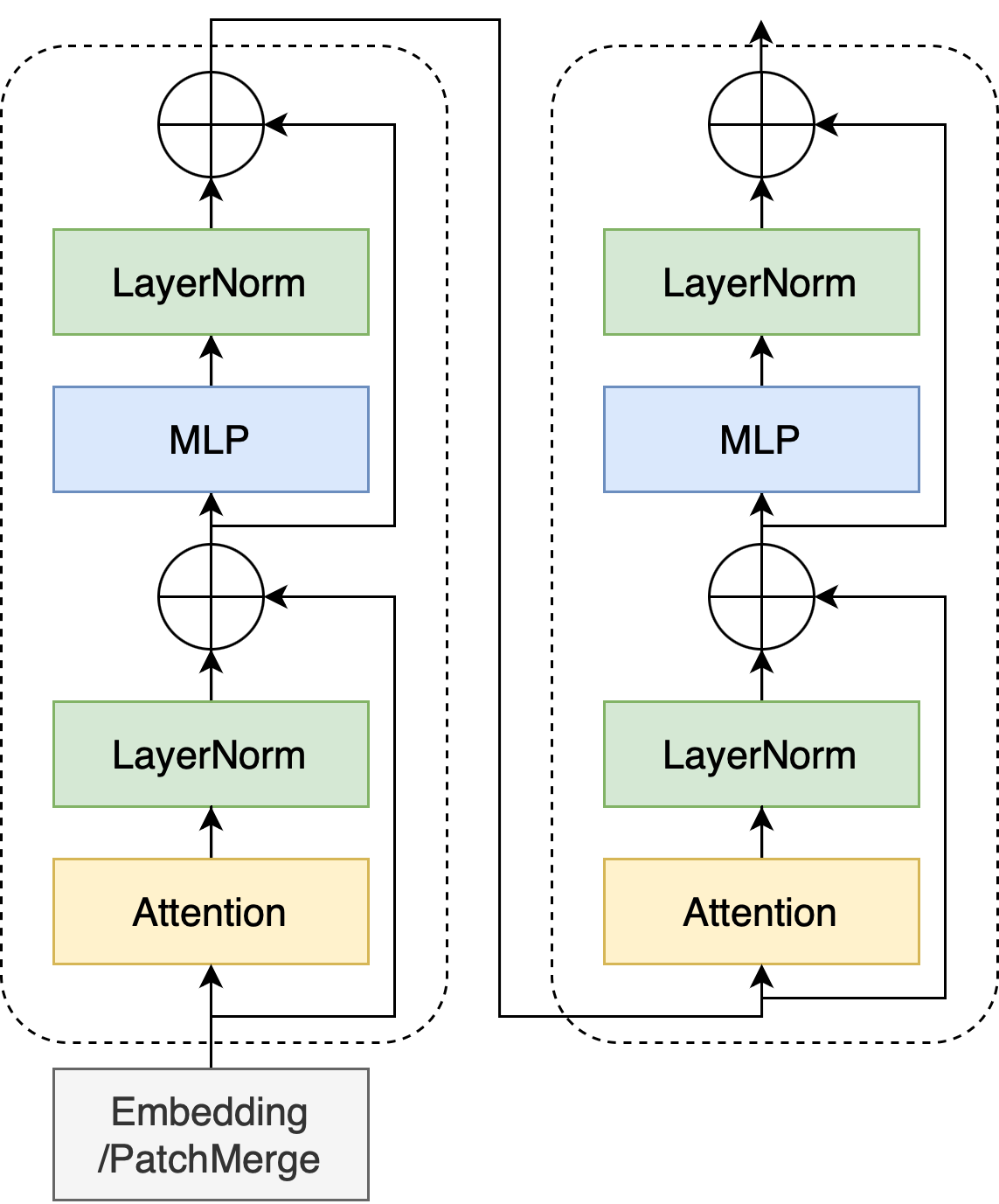}}
\caption{The Swin Transformer Block in the D-SarcNet model is made up of a linear embedding layer (for the first block) or a patch merging layer (for the other blocks), followed by two Transformer blocks.}
\label{fig_swin}
\end{figure}
Swin Transformer \cite{liu2021swin,liu2022swin}, standing for Shifted Window Transformer, was developed to address issues in various sectors, such as huge differences in the scale of visual components and high-resolution images. Similarly to ConvNeXt (as described in \ref{convnext}), Swin Transformer has four primary blocks. First, the model splits the image using the patch size of $4 \times 4$ into non-overlapping patches before applying it to the first block with a linear embedding layer to transfer raw-valued features into an arbitrary dimension and two successive Transformer blocks. After that, the patch merging layer is used to concatenate each group of $2 \times 2$ neighboring patches together as the input for the second block. This process is repeated four times to produce a hierarchical representation that uses the same ConvNeXt feature maps. Additionally, the multi-head self-attention (MSA) in the Transformer block is replaced by the shifted window-based MSA module to investigate the connections across windows. Swin Transformer also inherits visual prior of locality from the vanilla Transformer encoder, which is powerful in extracting local features from images. 


\subsubsection{Dual-stream ConvNeXt - Swin Transformer}
Since ConvNeXt and Swin Transformer employ different approaches to image feature extraction, this study introduces D-SarcNet, which integrates these two models into a dual-stream framework, as depicted in Fig. \ref{fig_framework}. The first stream, ConvNeXt, focuses on analyzing and extracting features directly from the raw hiPSC-CM images, each sized at $3 \times 224 \times 224$ (detailed in Section \ref{convnext}). While global information derived from raw images is crucial for sarcomere structure analysis, local information that captures diverse $\alpha$-actinin-2 patterns is equally important for assessing the maturity of hiPSC-CMs. To capture this local information, three distinct image-based representations — FFT Power, local patterns, and gradient magnitude — are proposed. These are generated by the FFT, a local-pattern model, and the Sobel operator, respectively. Specifically, FFT Power captures frequency-based features to identify periodic structures; local patterns specify multiple patterns within $\alpha$-actinin-2 structures; and gradient magnitude measures the rate of change in image intensities, effectively detecting the edges of z-discs and enhances $\alpha$-actinin-2 localization. For a detailed explanation of these processes, please refer to Appendix \ref{appendix:A}, \ref{appendix:B}, \ref{appendix:C}. These three image-based representations are resized to $224 \times 224$, stacked together, and then processed by the second stream, Swin Transformer (as described in Section \ref{swin}).

According to \cite{yao2024deep}, the global and local properties extracted from the two streams in terms of ConvNeXt and Swin Transformer should not be directly mixed and fed into the subsequent network. Therefore, the proposed framework analyzes these features separately and combines them at various scales. Specifically, after each stage, outputs are combined using addition operations and further processed by convolution layers ($3 \times 3$ and $1 \times 1$ with strides of one), batch normalization, ReLU activation, and max pooling as described in Fig. \ref{fig_framework}. The processed outputs from the four stages are concatenated to create a unified feature representation. Global averaging is subsequently applied to reduce the dimension of the concatenated features. Finally, to reduce overfitting, a succession of linear layers is used consecutively, followed by batch normalization.

\section{Experiments} \label{Experiments}

\subsection{Datasets and Experimental Settings}
\begin{figure}
\centerline{\includegraphics[width=0.5\textwidth]{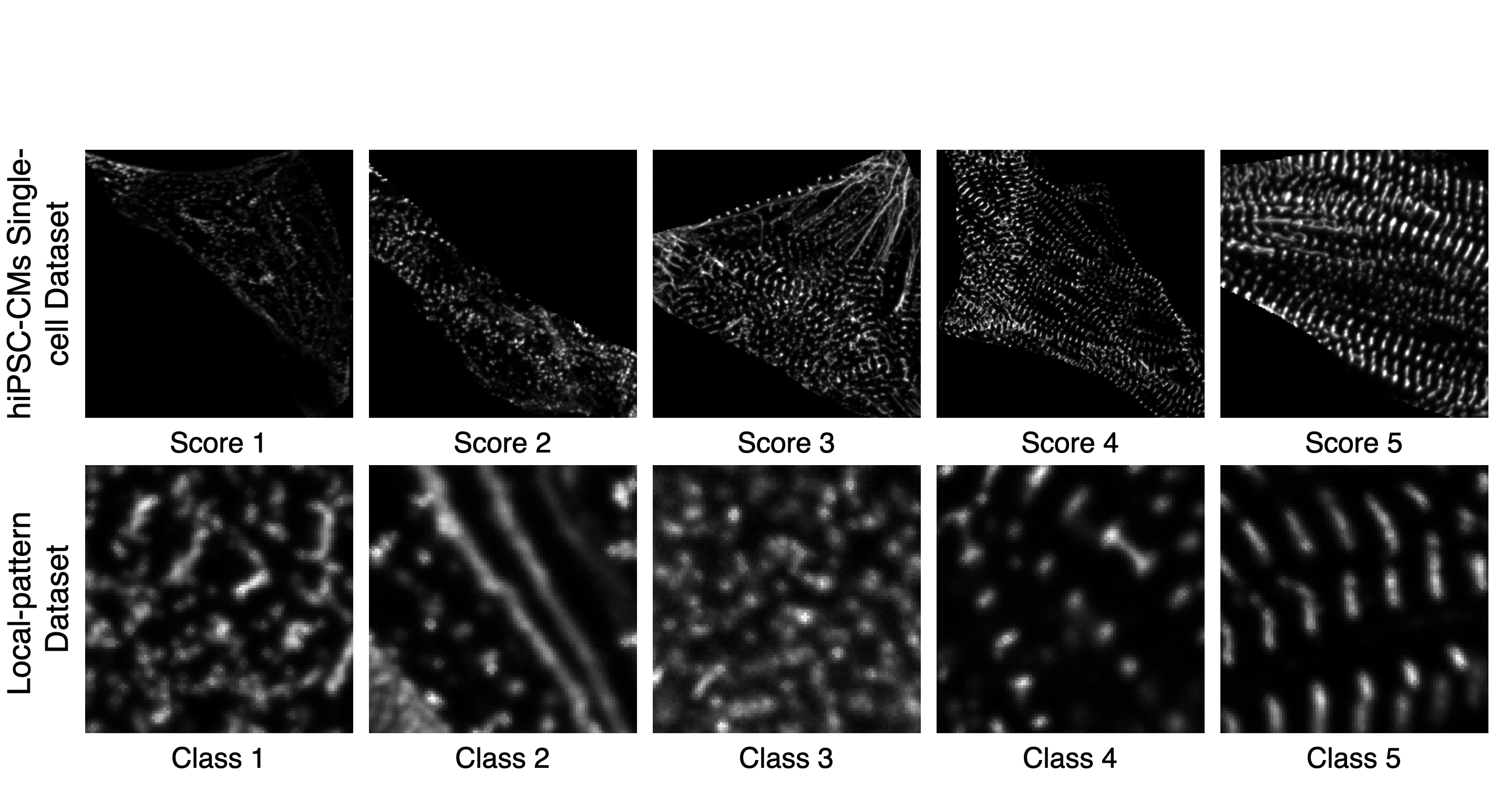}}
\caption{Representative examples of hiPSC-CMs single-cell images and local patterns for each category}
\label{fig_examples}
\end{figure}

Public datasets of fluorescent hiPSC-CMs at different stages are currently limited in number. Thus, this work uses only one openly available hiPSC-CMs single-cell dataset and the corresponding local-pattern dataset to train the D-SarcNet and local-pattern model, respectively. Fig. \ref{fig_framework} shows representative examples for both datasets. Details are provided below.

\subsubsection{hiPSC-CMs Single-cell Dataset} \label{dataset}
To verify the performance of the proposed approach, we conduct experiments on an open-source dataset of hiPSC-CMs single-cell images with the endogenously GFP-tagged$\alpha$-actinin-2 structure at days 18 and 32 from differentiation, provided by the Allen Institute for Cell Science (AICS) \cite{gerbin2021cell}. 

In this dataset, each cell undergoes manual scoring for the structural maturity of its sarcomere organization by two experts. The experts assign cells with five score groups from 1.0 to 5.0 based on the predominant organization of their structure. A score of 1.0 is given to cells with scattered, disordered puncta; a score of 2.0 is given to cells with more structured, denser puncta; a score of 3.0 is given to cells featuring both puncta and other types of structures such as fibers and z-discs; a score of 4.0 is given to cells with regular yet misaligned z-discs, and a score of 5.0 is given to cells with almost aligned z-discs. In our downstream analysis, we define the average score from the two experts as the ground truth. We also exclude all the cells with a score difference between the two experts greater than one from further analysis. After filtering, the final dataset contains 5,722 images of varying sizes. In particular, 81 cells are scored 1.0; 234 cells are scored 1.5; 428 cells are scored 2.0; 622 cells are scored 2.5; 2,541 cells are scored 3.0; 1,107 cells are scored 3.5; 548 cells are scored 4.0; 120 cells are scored 4.5; and 41 cells are scored 5.0. The dataset was split into 3,661 training images, 1,195 validation images, and 916 testing images for the experiments. 


\subsubsection{Local-pattern Dataset}
To train the local-pattern model, we used the dataset where the expert selects a subset of 18 representative examples from the hiPSC-CMs single-cell dataset described in Section \ref{dataset} and manually annotate 3,589 sub-regions within these images. Each sub-region is $96 \times 96$ pixels in size and is classified into one of five $\alpha$-actinin-2 pattern classes: diffuse/messy, fibers, disorganized puncta, organized puncta, and organized z-discs.

\subsection{Implementation Details and Evaluation Metrics}
\subsubsection{Implementation Details}
We first trained the local-pattern model and used that trained model to infer all single-cell images. The resulting inferences were then served as one of the inputs for the D-SarcNet model, referred to as local-pattern in Fig. ~\ref{fig_framework}. The images were resized to $224 \times 224$ pixels before being fed into the model. The network was built with Pytorch and trained on a GeForce RTX 3090 GPU. The batch size was set to 64. We used the Adam optimizer with a learning rate $1e-5$, training for 100 epochs. The seed is set to 1, and we implemented the same settings in all experiments for fair comparison.


\subsubsection{Evaluation Metrics}
To quantitatively evaluate the effectiveness of the proposed method, we used the following widely recognized regression metrics: Spearman
correlation, MAE, MSE, and the $R^2$ score.

First, Spearman correlation \cite{hollander2013nonparametric} measures the degree of association between two ranked variables. A perfect Spearman correlation of $+1$ or $-1$ means that one variable consistently increases or decreases in a perfectly predictable way as the other variable does the same. To compute this coefficient, firstly, $\hat{y}$ and $y$ are converted to ranks from lowest to highest. Let $d_i$ be the difference between the two ranks of each observation to calculate Spearman correlation as follows
\begin{eqnarray}
r_S = 1 - \frac{6 \sum_{i=1}^{N} d_i^2}{N(N^2 - 1)}.
\end{eqnarray}

Second, the Mean Absolute Error (MAE) computes the average absolute difference between $\hat{y}$ and $y$ as follows
\begin{eqnarray}
MAE = \frac{1}{N} \sum_{i=1}^{N} |y_i - \hat{y}_i|.
\end{eqnarray}

The Mean Squared Error (MSE) measures the average squared difference between $\hat{y}$ and $y$, computed as
\begin{eqnarray}
MSE = \frac{1}{N} \sum_{i=1}^{N} (y_i - \hat{y}_i)^2.
\end{eqnarray}

Lastly, the $R^2$ score, also known as the coefficient of determination \cite{wright1921correlation}, reveals how much of the variance in the dependent variable $\hat{y}$ that is predictable from the independent variable $y$. It varies from $-\infty$ to $+1$, with $+1$ being the optimal value. The $R^2$ score is calculated using the following formulation
\begin{eqnarray}
R^2 = 1 - \frac{\sum_{i=1}^{N} (y_i - \hat{y}_i)^2}{\sum_{i=1}^{N} (y_i - \overline{y})^2},
\end{eqnarray}
where $\overline{y}$ is the mean value of $y$.

\subsection{Experimental Results}
\subsubsection{Local-pattern Model}
We compare the performance of the local-pattern model, which uses the ConvNeXt framework, to the current state-of-the-art ResNet-18 module \cite{gerbin2021cell}. The experiments claim that the ConvNeXt framework has significantly enhanced the performance of local-pattern classification. Specifically, the ResNet-18 module achieves performance scores of 0.808, 0.811, 0.808, and 0.808 for the accuracy, precision, recall, and F1 score, respectively. Otherwise, the ConvNeXt framework outperforms the ResNet-18 module by approximately 5\% across these metrics, which are 0.854, 0.855, 0.854, and 0.853, respectively. This improvement highlights the substantial contribution and effectiveness of the ConvNeXt framework in local sarcomere pattern classification, leading to a significant advancement for the D-SarcNet model.

\subsubsection{D-SarcNet Model}
In this section, we compare the performance of the proposed method against: a) SarcNet \cite{le2024sarcnet}, the current state-of-the-art sarcomere structural organization scoring framework, and b) several state-of-the-art deep learning models with raw images as input. Table \ref{result} reports the experimental results of the D-SarcNet model and these frameworks. 

\begin{table}[htbp]
\caption{Experimental results on the testing dataset}
\begin{center}
\begin{tabular}{ | >{\centering\arraybackslash}m{11em} | >{\centering\arraybackslash}m{1.2cm} | >{\centering\arraybackslash}m{0.6cm} | >{\centering\arraybackslash}m{0.6cm} | >{\centering\arraybackslash}m{1.2cm} | } 
\hline
Method & Spear Corr ($r_S$) & MAE & MSE & $R^2$ Score  \\
\hline
SarcNet \cite{le2024sarcnet} & 0.831 & 0.310 & 0.161 & 0.668  \\
\hline
Swin Transformer \cite{liu2021swin} & 0.832 & 0.313 & 0.167 & 0.658  \\
\hline
DenseNet \cite{huang2017densely} & 0.813 & 0.340 & 0.194 & 0.599  \\
\hline
ResNet-50 \cite{he2016deep} & 0.849 & 0.286 & 0.138 & 0.713  \\
\hline
ConvNeXt \cite{liu2022convnet} & 0.856 & 0.278 & 0.128 & 0.733  \\
\hline
\textbf{D-SarcNet (ours)} & \textbf{0.868} & \textbf{0.265} & \textbf{0.119} & \textbf{0.753}  \\
\hline
\end{tabular}
\label{result}
\end{center}
\end{table}

Experiments on the hiPSC-CMs single-cell dataset show that the proposed approach significantly outperforms SarcNet and the four state-of-the-art deep learning models in all performance metrics. In particular, on the testing set, D-SarcNet reports a Spearman correlation of 0.868, an MAE of 0.265, an MSE of 0.119, and a $R^2$ score of 0.753. We observe that these results are much better than the performance of the SarcNet framework with 3.7\%, 4.5\%, 4.2\%, and 8.5\% improvement in Spearman correlation, MAE, MSE, and $R^2$ score, respectively. 

In addtion, D-SarcNet surpasses all other state-of-the-art models, including Swin Transformer \cite{liu2021swin}, DenseNet \cite{huang2017densely}, ResNet-50 \cite{he2016deep}, and ConvNeXt \cite{liu2022convnet}. For instance, compared to ConvNeXt, D-SarcNet shows a 1.2\% improvement in Spearman correlation, a decrease of 1.3\% in MAE, a 0.9\% decrease in MSE, and a 2\% increase in the $R^2$ score. From the table, we can also notice that ConvNeXt achieves the best scores among all four deep-learning models. This indicates that ConvNeXt baseline is the best model to classify sarcomere structure organization from raw images.

\subsubsection{Ablation studies}
To validate the effectiveness of D-SarcNet and assess its ability to utilize information from all three image-based representations effectively, we evaluate the performance of the D-SarcNet in four scenarios: (1) removal of one stream, (2) without the blocks-combined components, (3) without post-processing after combining blocks from each stream, and (4) using single image-based representation. Table \ref{abalation} reports the performance of the main indicators in each experiment.


\begin{table*}[htbp]
\caption{Ablation experimental results on the testing dataset}
\begin{center}
\begin{tabular}{|l|l|c|c|c|c|}
\hline
\multicolumn{1}{|c|}{Stream} & \multicolumn{1}{c|}{Method}                                         & Spear Corr ($r_S$)     & MAE            & MSE            & $R^2$ Score       \\ \hline
\multirow{2}{*}{Single}      & ConvNeXt stream (raw images)                                                           & 0.859          & 0.272          & 0.125          & 0.739          \\ \cline{2-6} 
                             & Swin Transformer stream (FFT Power, Local patterns, Gradient magnitude)                                                  & 0.830                &0.302                &0.150                &0.687                \\ \hline
\multirow{6}{*}{Dual}        & D-SarcNet (without blocks-combined)                                & 0.860          & 0.265          & 0.127          & 0.735          \\ \cline{2-6} 
                             & D-SarcNet (without post-processing)                                & 0.865          & 0.268          & 0.123          & 0.743          \\ \cline{2-6} 
                             & D-SarcNet (Only FFT Power on the Swin-Transformer stream)                & 0.862          & 0.266          & 0.122          & 0.745          \\ \cline{2-6} 
                             & D-SarcNet (Only Local patterns on the Swin-Transformer stream)     & 0.864          & \textbf{0.264} & 0.123          & 0.744          \\ \cline{2-6} 
                             & D-SarcNet (Only Gradient magnitude on the Swin-Transformer stream) & 0.863          & 0.268          & 0.125          & 0.740          \\ \cline{2-6} 
                             & \textbf{D-SarcNet (Ours)}                                                   & \textbf{0.868} & 0.265          & \textbf{0.119} & \textbf{0.753} \\ \hline
\end{tabular}
\label{abalation}
\end{center}
\end{table*}

First, the dual ConvNeXt-Swin Transformer framework significantly outperforms each single stream. Using the ConvNeXt stream alone, which processes raw images to extract global features, results in a Spearman correlation of 0.859 (a decrease of 0.9\%), an MAE of 0.272 (an increase of 0.7\%), an MSE of 0.125 (an increase of 0.6\%), and an $R^2$ score of 0.739 (a decrease of 1.4\%). The Swin Transformer stream, which analyzes FFT Power, local patterns, and gradient magnitude to capture local features results in decrease of 3.8\% in Spearman correlation, an increase of 3.7\% in MAE, an increase of 3.1\% in MSE, and a decrease of 6.6\% in $R^2$ score of 0.687. These results show that combining ConvNeXt and Swin Transformer with blocks-combined components provides more robust feature extraction at both global and local levels.

Second, five ablation experiments in Dual-stream confirm the contribution of each component in the proposed framework. The model without blocks-combined (D-SarcNet without blocks-combined) shows lower performance than the full framework, results in a decrease of 1.8\% in $R^2$ score and an increase of 0.8\% in MSE. This demonstrates that blocks-combined components facilitate better generalization and more precise predictions. Similarly, removing post-processing after combining blocks from each stream (D-SarcNet without post-processing) also results in a performance drop, with the Spearman correlation decreasing to 0.865, MAE increasing to 0.268, MSE increasing to 0.123, and $R^2$ score decreasing to 0.743. This highlights that the post-processing steps add value to the overall performance by ensuring that the synthetic features from each stage of the two streams are optimally combined and normalized via different sizes of receptive fields.

When replacing the input in the second stream with each image-based representation individually, we observe a decrease in performance. However, these models still outperform the single-stream ConvNeXt model, indicating their contribution to the overall framework. The framework using FFT Power achieves a Spearman correlation of 0.862 and an $R^2$ score of 0.745, demonstrating its effectiveness in capturing frequency-based features of sarcomere structures. The framework with local-pattern representation shows slightly better performance with a Spearman correlation of 0.864, the lowest MAE of 0.264, and an $R^2$ score of 0.744, emphasizing its contribution by providing information on the diversity of sarcomere structural patterns. The framework with gradient magnitude representation achieves a Spearman correlation of 0.863 and an $R^2$ score of 0.740, capturing information on the edges of z-discs and $\alpha$-actinin-2 visualization. These results confirm that each image-based representation provides unique and complementary information for analyzing the complex internal structure of sarcomere organizations in hiPSC-CMs. 

\subsection{Quantitative Measurement of Sarcomere Properties}

\begin{figure*}
\centerline{\includegraphics[width=0.7\textwidth]{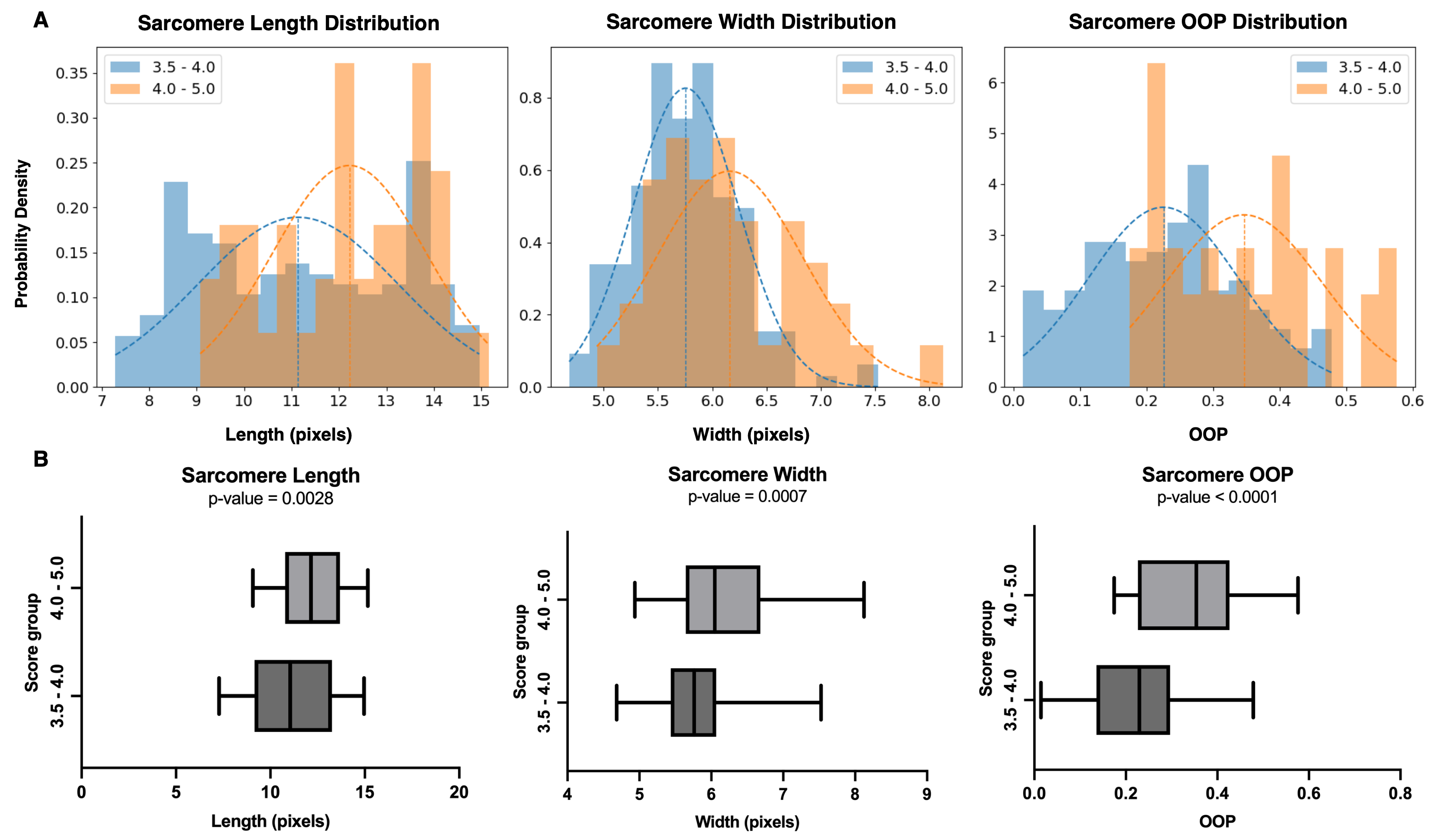}}
\caption{Quantitative measurement of sarcomere properties, including sarcomere length, sarcomere width, and OOP, on the two groups: the first group with predicted scores from 3.5 to 4.0 and the second group with predicted scores from 4.0 to 5.0. (A) Histograms on the first group (blue) and the second group (orange). (B) Box plots and the p-values are represented above the box plots on the two groups.}
\label{fig_histogram}
\end{figure*}

As mature hiPSC-CMs are longer and exhibit a higher level of structural organization compared to immature ones, this section aims to confirm the reliability of the predicted results using quantitative measurement of sarcomere properties, including sarcomere length, sarcomere width, and orientational order parameter (OOP). These three metrics are calculated by the SarcGraph algorithm \cite{zhao2021sarc}, in which the z-discs are segmented and two adjacent z-discs are paired up with each other. The problem is the potential bias when the algorithm only works well on segmenting well-formed sarcomere structures. As a result, in this section, we only run the SarcGraph algorithm on the images with predicted results equal to or larger than 3.5, indicating almost organized z-discs. Fig. \ref{fig_histogram} compares the differences in the three metrics mentioned above among two groups: (1) scores from 3.5 to 4.0 and (2) scores from 4.0 to 5.0. It is noticeable that the three figures in the second group are significantly larger than the ones of the first group ($p=0.0028$ for sarcomere length, $p=0.0007$ for sarcomere width, and $p<0.0001$ for OOP). OOP should be noted as the most remarkable measurement when $p<0.0001$ and the range of histogram shifts by approximately 0.18.

\section{Discussion \& Conclusion}
\label{Discussion and Conclusion}
In this work, we propose D-SarcNet, a deep learning framework to automatically score sarcomere organization in fluorescently labeled hiPSC-CMs single-cell images, outperforming the current state-of-the-art without requiring the prior feature engineering process. We design a dual-stream framework architecture combining ConvNeXt and Swin Transformer for global and local feature extraction. Remarkably, we propose using FFT Power, deep learning-generated local patterns, and gradient magnitude as input for the second stream to provide the framework more information on the heterogeneity in sarcomeric organizational states. Extensive experiments and ablation studies demonstrate the advantages of the proposed framework over existing state-of-the-art methods and confirm the contribution of the proposed image-based representations in sarcomere analysis. The lack of progress in single-cell segmentation remains a limitation. In the AISC dataset \cite{gerbin2021cell}, experts manually draw single-cell boundaries due to the lack of an accessible automated framework. This work should be deemed difficult because the single cells are overlap. Future work will focus on developing the hiPSC-CMs single-cell segmentation framework and exploring the potential of this framework in cardiac research pipeline.

\bibliographystyle{IEEEtran}
\bibliography{references}

\appendix




This section provides more details on the image-based transformation techniques we used to train the Swin Transformer stream, including FFT, local-pattern model and Sobel operator. 

\subsection{FFT}
\label{appendix:A}
FFT is a method that effectively computes the Discrete Fourier Transform (DFT) by leveraging symmetries, which are maximized when the number of points $n$ is a power of two \cite{cooley1965algorithm}. In hiPSC-CM analysis, FFT is also utilized to monitor changes in sarcomere length and evaluate homogeneous populations of cardiomyocytes in linearly aligned sarcomeres \cite{toepfer2019sarctrack}. In this study, the raw image $x^{(i)}$ would be divided into multiple windows $w^{(j, k)}$ with the size of $96 \times 96$ and the step of eight. Let defined $w^{(j, k)}(0, 0), ..., w^{(j, k)}(n-1, n-1)$ be the data points of the given input $w^{(j, k)}$ with size $n \times n$. The formula for two-dimensional input $w^{(j, k)}$ would be given as
\begin{equation}
X(u, v) = \sum_{p=0}^{n-1}  \sum_{q=0}^{n-1} w^{(j, k)} (p, q) \times e^{-2 \pi i (\frac{up}{n} + \frac{vq}{n})},
\end{equation}
where $X(u,v)$ is the DFT at the frequency coordinates $(u,v)$, $w^{(j, k)}(p,q)$ is the value of the image at the spatial coordinates $(p,q)$, and $i$ is the imaginary unit. 

The FFT Power image, which reveals the periodicity within the signal by displaying peaks at the corresponding frequencies, would then be calculated as
\begin{equation}
P(j, k) = \sum_{u=0}^{n-1}\sum_{v=0}^{n-1}|X(u, v)| ^ 2,
\end{equation}
where $P(j, k)$ is the value of FFT Power corresponding to the window $w^{(j, k)}$ at coordinates $(j, k)$ of the FFT Power image.

\subsection{Local-pattern Model}
\label{appendix:B}
The local-pattern mode classifies the local organization of $\alpha$-actinin-2 pattern in hiPSC-CM images into five classes: diffuse/messy, fibers, disorganized puncta, organized puncta, and organized z-discs. The training process is described in Section \ref{dataset}. Specifically, the patches of size $96 \times 96$ pixels are extracted and centered around each labeled point. These patches are then interpolated into $224 \times 224$ and serve as input for the ConvNeXt model (as depicted in Section \ref{convnext}). 

During the inference phase on single-cell hiPSC-CMs images, the algorithm divides each image using a step size of eight pixels into overlapping windows with the size of $96 \times 96$. For each $96 \times 96$ window, the trained model outputs a class between 1 and 5. These values are then mapped to their corresponding locations in the raw hiPSC-CM image to create a maturity map. The background regions are marked as zeros.

\subsection{Sobel Operator}
\label{appendix:C}
The Sobel operator, a widely used gradient operator in image processing for edge detection \cite{sobel1990isotropic}, is applied in this study to measure intensity changes around pixels, enabling edge detection of z-discs, enhancing $\alpha$-actinin-2 patterns, and reducing noise. The gradient at a point $f(x, y)$ is computed using the central difference method over a $3 \times 3$ neighborhood in the $x$ and $y$ directions. The convolution templates are
\begin{equation}
G_x =
\begin{bmatrix}
    -1 & 0 & 1 \\
    -2 & 0 & 2 \\
    -1 & 0 & 1
\end{bmatrix},
G_y =
\begin{bmatrix}
    -1 & -2 & -1 \\
    0 & 0 & 0 \\
    1 & 2 & 1
\end{bmatrix},
\end{equation}
where $G_x$ is the horizontal kernel, $G_y$ is the vertical kernel.

After that, the gradient magnitude would be calculated as
\begin{equation}
G(x, y) = \sqrt{G_x^2 + G_y^2},
\end{equation}
where $G(x, y)$ represents the gradient magnitude at the raw image coordinates $(x, y)$.

\end{document}